\documentclass[a4paper,twoside]{article}

\usepackage{epsfig}
\usepackage{subcaption}

\usepackage{calc}
\usepackage{amssymb}
\usepackage{amstext}
\usepackage{amsmath}
\usepackage{amsthm}
\usepackage{multicol}
\usepackage{pslatex}
\usepackage{apalike}
\usepackage{hyperref}
\usepackage{url}

\usepackage{SCITEPRESS}     

\begin{document}

\title{Network of Steel: Neural Font Style Transfer from Heavy Metal to Corporate Logos}

\author{\authorname{Aram Ter-Sarkisov}
\affiliation{City, University of London}
\affiliation{Artificial Intelligence Research Centre, CitAI}
\email{alex.ter-sarkisov@city.ac.uk}
}


\abstract{We introduce a method for transferring style from the logos of heavy metal bands onto corporate logos using a VGG16 network. We establish the contribution of different layers and loss coefficients to the learning of style, minimization of artefacts and maintenance of readability of corporate logos. We find layers and loss coefficients that produce a good tradeoff between heavy metal style and corporate logo readability. This is the first step both towards sparse font style transfer and corporate logo decoration using generative networks. Heavy metal and corporate logos are very different artistically, in the way they emphasize emotions and readability, therefore training a model to fuse the two is an interesting problem.}

\onecolumn \maketitle \normalsize \setcounter{footnote}{0} \vfill

\section{\uppercase{Introduction}}
\label{sec:introduction}

\noindent Recently there has been a large number of applications of  convolutional neural networks (ConvNets) to neural style transfer. VGG16 \cite{simonyan2014very} was used to extract features from both content and style images \cite{gatys2016image} to transfer style onto a randomly created image or the content image. This approach was improved in \cite{zhang2017multi} by adding upsampling layers and making the network fully convolutional. A number of generative adversarial networks, GANs \cite{goodfellow2014generative} were developed and successfully applied to the neural style transfer for images and videos, such as CycleGANs \cite{zhu2017unpaired}, Pix2pix \cite{isola2017image}, pose-guided GANs \cite{ma2017pose}.\\

\noindent Font neural style transfer is an area of neural style transfer that is concerned with the transfer and generation of font styles. In \cite{azadi2018multi} GAN was developed that synthesizes unseen glyphs (characters) given the previously observed ones in a particular decorative style. In \cite{yang2019controllable} GANs are trained to transfer style (fire, water, smoke) to glyphs to create an artistic representation. GlyphGAN \cite{hayashi2019glyphgan} was recently developed for generation of glyphs in a required style. Neural font transfer for logo generation \cite{atarsaikhan2018contained} uses a framework similar to \cite{gatys2016image}, i.e. minimizes distance to the style and content images by extracting features from ConvNet (VGG16) layers.\\

\noindent In this publication we would like to extend these findings to a sparse case of logo style transfer: from the style image, logo of a heavy metal band we want to extract only foreground font ignoring the background. Content images are corporate logos. To the best of our knowledge this is the first attempt to train such a model. In Section \ref{sec:model} we introduce Network of Steel that learns to transfer heavy metal logo style while maintaining corporate logo structure, Section \ref{sec:exps} presents the main results of the experiments, Section \ref{sec:conclusion} concludes.   

\section{\uppercase{Our Approach}}
\label{sec:model}
\noindent We introduce Network of Steel that learns to transfer the style from the heavy metal logo while maintaining the structure of the corporate logo. We compare two models, one based solely on VGG16 \cite{gatys2016image} and the other on Multistyle Generative Network \cite{zhang2017multi}. The advantage of the former is that it does not require a large dataset for training; instead, only one content and one style image are used. 
\subsection{Heavy metal and corporate logo style}
\begin{figure*}
    \centering
     \begin{subfigure}{.5\textwidth}
     \centering
     \includegraphics[width=\linewidth]{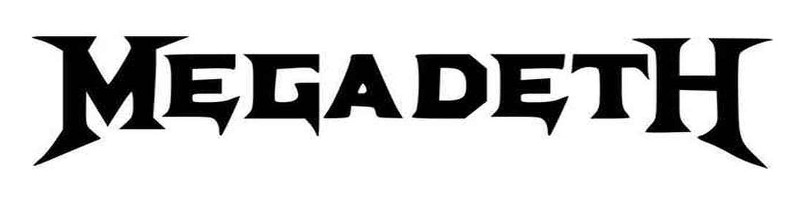}
     \caption{Megadeth logo, black and white}
          \label{fig:megadeth}
      \end{subfigure}\hfill
     \begin{subfigure}{.5\textwidth}
     \centering
     \includegraphics[width=\linewidth]{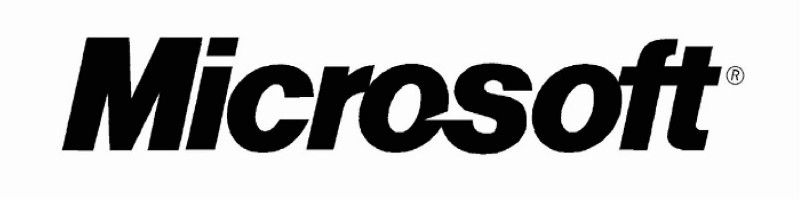}
     \caption{Microsoft logo, black and white}
          \label{fig:microsoft}
     \end{subfigure}
    \caption{Examples of heavy metal and corporate logos}
    \label{fig:logos}
\end{figure*}

\noindent In this publication we only concern ourselves with the style of band logos, leaving out the style of album covers, which is an entirely different area. Logos of heavy metal bands are usually carefully designed in order to convey a certain group of emotions or messages, usually those of fear, despair, aggression, alertness, eeriness, mystery. These logos are often a true work of art. Several features stand out in many logos: the first and the last glyphs are more prominent, often elongated and symmetric around the center of the logo, most glyphs are decorated in the same style, e.g. Megadeth logo glyphs have sharpened kinks at the edges (arms), the first glyph ($M$) and the last glyph ($h$) are about twice the size of other glyphs, with extrusions and slanted bottom kinks. The logo is also symmetric around the horizontal and vertical axes, see Figure \ref{fig:megadeth}. \\

\noindent On the other hand, corporate logos are often barely distinguishable from plain text. Their design, although often expensive in development, tends to be functional, vapid and boring, with an emphasis on readability and recognizability, see Figure \ref{fig:microsoft} for Microsoft logo. This publications intends to bridge the gap between the two by transferring the style from the heavy metal band (Megadeth) to a corporate logo (Microsoft).\\

\noindent Heavy metal band logos are an example of sparse style in a sense that we only want to learn and transfer font (glyph) features keeping the corporate logo's white background. This often leads to the creation of a large number of artefacts, such as color pixels. 

\subsection{VGG16 and MSG Net} 
\begin{figure*}
  \centering
   \includegraphics[scale=0.5]{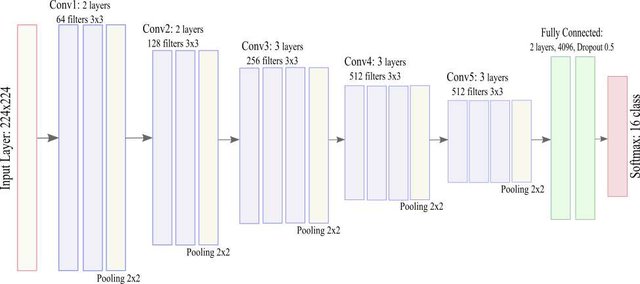}
  \caption{VGG16 architecture. There are in total five blocks, the first two blocks have two \texttt{Conv} layers, each followed by \texttt{ReLU} and \texttt{MaxPool} layers, the last three have three \texttt{Conv} layers, each followed by \texttt{ReLU} and \texttt{MaxPool} layers. Image taken from \cite{das2018document}.}
  \label{fig:vgg16}
 \end{figure*}

\noindent We compare two models, VGG16 \cite{simonyan2014very} used by \cite{gatys2016image} to transfer style and multi-style generative network \cite{zhang2017multi}, which uses Siamese network to extract features from the style image, fully convolutional network (FCN) to extract features from the content image and co-match layer to find correlation. VGG16 is presented in Figure \ref{fig:vgg16}
In this publication we refer to the relevant \texttt{Conv} (convolution) layer using the style adapted in most deep learning publications and code,\texttt{convi${}\_{}$j} , where \texttt{i} refers to the block in VGG16 and \texttt{j} to the \texttt{Conv} layer in that block. Block in VGG16 is an informal, but very useful term for our purpose. The first two blocks have 2 \texttt{Conv} layers each, with 64 and 128 feature maps, equipped with \texttt{ReLU} and \texttt{MaxPool} layers. The next three blocks have 3 \texttt{Conv} layers each, also followed by \texttt{ReLU} and \texttt{MaxPool} layers, with 256, 512 and 512 feature maps.
\subsection{Network of Steel}
Following the framework of \cite{gatys2016image}, we use exactly one style image, one content image and one synthesized image that is updated every iteration of the algorithm. Content and style features  are extracted once before the start of the algorithm, and features extracted from the synthesized image are compared to them every iteration to obtain the loss value, backpropagated it and update the synthesized image.\\

\noindent Our model, which we refer to as Network of Steel, is VGG16 without the classifier layers (\texttt{fc6,fc7}). Most generative networks use one last \texttt{ReLU} layer from every block for style features extraction and the second \texttt{ReLU} from the fourth block for content loss \cite{gatys2016image}. Our first contribution is the use of \texttt{Conv} layers for feature extraction and loss computation, because \texttt{ReLU} layers produce multiple artefacts. Our second contribution is the use of coarse features from the first block (\texttt{conv1${}\_{}$1}, \texttt{conv1${}\_{}$2}).\\

\noindent We show that extracting coarse features from the style image minimizes artefacts in the stylized logo and it is sufficient to use only two deep layers in addition to a coarse layer to transfer style without distorting the corporate logo or creating many artefacts. Finally, our third contribution are the layerwise loss weights that determine the contribution of the layer to the style loss. We show that VGG16 outperforms MSG Net for style transfer. \\

\noindent The specific challenge that Network of Steel faces is the sparse distribution of style features: the network only needs to learn to transfer font features (shapes, ornaments, glyph sizes and ratio, overall symmetry) to the synthesized image and keep the background neutral.\\      

\noindent To find the content loss, we use the same layer as in \cite{gatys2016image}, \texttt{conv4${}\_{}$2}, and minimize mean squared error function between the features of the content and the synthesized image. Each layerwise style loss is multiplied by the predefined loss coefficient; if the coefficient is different from $0$, we refer to the corresponding layer as an $active$ layer:
\begin{equation}\label{eq1}
    \text{L}_{Total} = \text{L}_{Content} + \sum_{l=0}^{L}c_l E^l
\end{equation}
Coefficients $c_l$ are layerwise loss coefficients specified in Section \ref{sec:exps} and Appendix. For example, for layer \texttt{conv3${}\_{}$3} layer loss coefficient is \texttt{c$_{3\_3}$}. For style and transferred image we compute correlation matrices $\mathbf{A^l}$ and $\mathbf{G^l}$ for every active layer. Each element of $\mathbf{G^l}, G^{l}_{ij}$ is a dot-product of vectorized feature maps in that layer:
\begin{equation}
\label{eq2}
    G^{l}_{ij} =\mathbf{f^{T}_i} \cdot  \mathbf{f_j}
\end{equation}
This array is known as Gram matrix. Distance between Gram matrices for the style and synthesized images, $\mathbf{A^l}$ and $\mathbf{G^l}$ is a contribution to the style loss, and it is measured using mean squared error:
\begin{equation}
\label{eq3}
   E^l = \frac{1}{4 H^2 W^2} \sum_{i,j}^{C^2} (A^l_{ij} - G^l_{ij})^2
\end{equation}
Equation \ref{eq1} is different from the total loss equation in \cite{gatys2016image} because we only use layer-specific style loss coefficients, and do not use the style loss coefficient. Here $H$ is the height of the feature map, $L$ is the length of the feature maps and $C$ is the number of channels/feature maps in layer $l$. We set most layerwise coefficients to $0$, and establish which active layers contribute most to the creation of a readable corporate logo with the best heavy metal style features and least artefacts, which is the main contribution of this publication. 
\section{\uppercase{Experiments}}
\label{sec:exps}
\noindent We compare the results of Network of Steel, MSG Net and VGG16 as implemented in the GitHub repository  \url{https://github.com/rrmina/neural-style-pytorch}. VGG16 uses \texttt{Conv} layers \texttt{conv1${}\_{}$2, conv2${}\_{}$2, conv3${}\_{}$3, conv4${}\_{}$3, conv5${}\_{}$3}. \\

\noindent Weights in VGG16 which extracts features from the synthesized image are frozen for the full duration of the algorithm, and only partial derivatives for the pixels \textbf{x} in the synthesized image are computed and updated: $\frac{\partial L_{Total}}{\partial \mathbf{x}} \neq 0$.\\

\noindent Network of Steel is always initialized with the content image and pretrained VGG16 weights. Only specified layers contribute to the style loss. For training we use Adam optimizer with a learning rate of $1.0$ and regularization constant of $0.001$. We use the same MSG-Net hyperparameters as in the GitHub project: \url{https://github.com/zhanghang1989/MSG-Net}.\\

\noindent We select three values for layerwise coefficients: $0, 20, 200, 2000$. If the coefficient is set to $0$, the style loss from this layer is not computed and ignored during training. Only losses from the active layers contribute to the total loss. To test our ideas, we tried three different groups of layers to compute style loss:
\begin{enumerate}
    \item Two layers, the fist one is always \texttt{conv1${}\_{}$2}, the other layer is the last \texttt{Conv} layer from one of the last three blocks. 
    \item Three layers, the fist one is always \texttt{conv1${}\_{}$2}, the other two layers are the two last \texttt{Conv} layers from one of the last three blocks. 
    \item Two blocks, the fist one is always the first block of VGG16, the other one is one of the last three blocks. 
\end{enumerate}

Obviously far more combinations of layers and loss coefficients can be tested, but empirically we determined that adding more layers and increasing loss coefficients leads to the deterioration of the network's performance, so these results are not presented. 

\subsection{Baseline VGG16 model}
Results for the generative network from \cite{gatys2016image} are presented in Table \ref{tab:gatysmodel}. In the original publication and code the last \texttt{ReLU} layer of each block is used with different weights, but most open-source implementations use total style loss of 1e2 and layerwise loss coefficient of $0.2$. We use a higher style loss weight $=1e5$. In addition to $0.2$ we try coefficients of $0.02$ and $0.002$. For the lowest loss coefficent, the model displays artefacts (color pixels). For higher coefficients it distorts the font and makes the logo unreadable. This demonstrates that the selection of layers for loss computation alone is not enough for sparse style transfer, and adjusting loss coefficients affects the presence of artefacts and degree of style transfer. 
\subsection{Two style layers}
Content features include the structure (text) of the corporate logo, but learning these features always leads to a large number of artefacts in the transferred image. To minimize these artefacts, the second \texttt{conv} layer from the first block, \texttt{conv1${}\_{}$2} is used. This coarse layer is responsible for the learning of color features from the style logo, which transfers the background features (white color), but does not learn font features like the edgy shapes of the glyphs in Megadeth logo. Using only one other layer from the last three blocks increases both the effect of style glyph features, including the elongation of the first and the last glyphs (elongated bottom of the front `leg' of $M$ and $t$, as particularly obvious in Tables \ref{tab:c12c43} and \ref{tab:c12c33}, and the presence of the artefacts: font decay, colored pixels in the background, wrong style features (like a vertical `extension' to glyph `o' in Table \ref{tab:c12c33}). \\

\noindent Increasing loss coefficient for the deep layers always leads to the transfer of more subtle features, like the more `rectangular' shape of the intermediate glyphs (all glyphs except the first and the last one). More sophisticated style features, like small horizontal extrusions on top of `legs' of $M$ and $t$ remain a challenge with two style layers. Either the model evolves small black vertical `blob', like in Table \ref{tab:c12c53}, or merges this extrusion with the dot over $i$, the next glyph. For the same glyph $M$, the elongated bottom of the front leg is a lesser challenge, (see Table \ref{tab:c12c33} with \texttt{conv3${}\_{}$3}, to a lesser extent this is achieved in Table \ref{tab:c12c43} with \texttt{conv4${}\_{}$3}). \\

\noindent It is noticeable that the last layer in the network, \texttt{conv5${}\_{}$3} contributes very little to the style transfer. We explain this by the size of the layer and the size of the loss it generates. \texttt{conv4${}\_{}$3} and \texttt{conv3${}\_{}$3} have either the same (512) or fewer (256) number of maps, but they are larger. Convergence of the networks with the largest coefficients, \texttt{conv1${}\_{}$2} = \texttt{conv5${}\_{}$3} = \texttt{conv4${}\_{}$3} = \texttt{conv3${}\_{}$3}=2000, is shown in Figure \ref{fig:conv1}. For the same size of the loss coefficient, \texttt{conv3${}\_{}$3}'s contribution to the style loss is much higher than of the other layers.



\subsection{Three style layers}
The results for the fifth block are very similar to the previous case with two style layers: the network fails to learn any significant style features. Same is true for the fourth block, only when the largest loss coefficient is used it manages to evolve some style features, like the elongation of the last glyph. The third block learns the style very aggressively, and for \texttt{c$_{3\_2}$}=\texttt{c$_{3\_3}$}=200 most style features are learnt, including the horizontal kinks on top of the first glyph, but at the cost of overall deterioration of the logo quality, see Table \ref{tab:c12c32c33}.\\

\noindent This is reflected in the convergence of the networks, see Figure \ref{fig:conv2}. Despite some progress compared to the networks with two layers, there seems to be a lack of good tradeoff between learning of style and maintaining the structure of the corporate logo. This is reflected in the addition of another coarse layer in the next subsection.   
\subsection{Two style blocks} 
\noindent With two blocks, we use all the \texttt{Conv} layers in each block. Empirically we established that adding even more layers or whole blocks either does not improve the result, or the logo readability starts to deteriorate.\\

\noindent To find a better tradeoff between readability and style transfer, we add another coarse layer, \texttt{conv1${}\_{}$1}, so the total number of active layers increased to five, same as in \cite{gatys2016image}. This effect explains why some results have fewer style features than in the previous subsection, despite adding another deep layer.\\ 

\noindent The overall result for the fifth block in Table \ref{tab:c11c12c51c52c53} is still quite weak, but with the largest loss coefficients for all layers it manages to transfer some style for the first and the last glyphs, and the intermediary glyphs, with very few artefacts. The third block contributes much more to the style loss, so the best results is achieved for \texttt{c$_{1\_1}$}=\texttt{c$_{1\_2}$}=2000 and \texttt{c$_{3\_1}$}=\texttt{c$_{3\_2}$}=\texttt{c$_{3\_3}$}=20, but further increase of style loss coefficients leads to the deterioration of the transferred logo. Nevertheless, for the largest loss coefficients the network almost correctly evolves the first and the last glyphs at the cost of adding few background artefacts and slight deterioration of the readability of the synthesized logo.\\

\begin{figure*}
    \centering
     \begin{subfigure}{.33\textwidth}
     \centering
     \includegraphics[width=\linewidth]{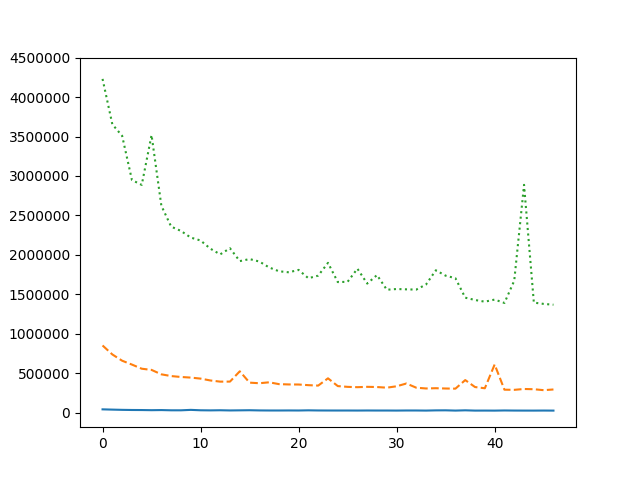}
     \caption{Two style layers} 
          \label{fig:conv1}
\end{subfigure}\hfill
     \begin{subfigure}{.33\textwidth}
     \centering
     \includegraphics[width=\linewidth]{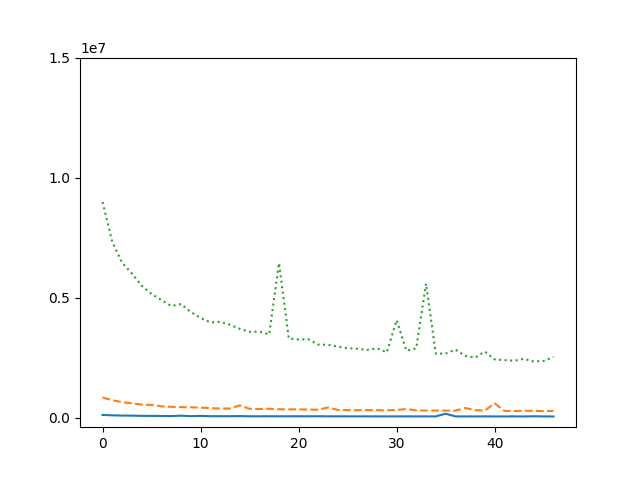}
     \caption{Three style layers}
     \label{fig:conv2}
     \end{subfigure}\hfill
     \begin{subfigure}{.33\textwidth}
     \centering
     \includegraphics[width=\linewidth]{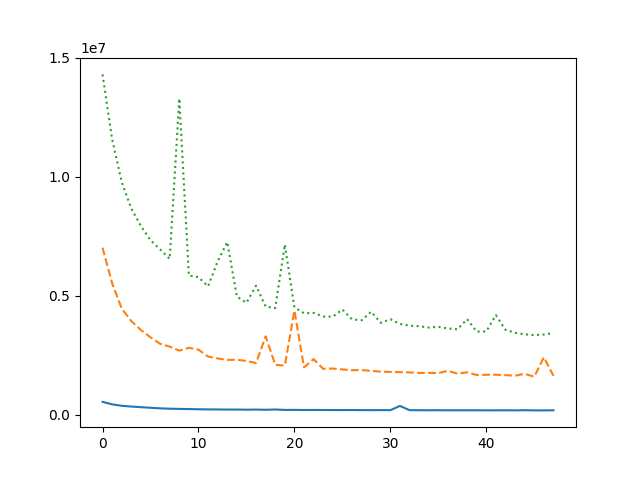}
     \caption{Two style blocks}
      \label{fig:conv3}

     \end{subfigure}\hfill
    \caption{Style loss for each network with all loss coefficients $=2000$ plotted against thousands of iterations. Continuous curve: fifth block layers, dashed curve: fourth block layers, dotted curve: third block layers.}
    \label{fig:loss_convergence}
\end{figure*}

\noindent The best results that balance the heavy metal style and corporate logo readability were obtained using the fourth block with the style loss coefficients of $200$ and  \texttt{c$_{1\_1}$}=\texttt{c$_{1\_2}$}$=2000$ in Table \ref{tab:c11c12c41c42c43}: the model evolved most of the challenging features of the metal logo without a heavy presence of artefacts. They are an improvement over the results in Table \ref{tab:c12c42c43} in most ways, which proves that adding both a deep style layer and a coarse layer to maintain the content font structure improves the final result. This could be seen in Figure \ref{fig:conv3}: the third block generates more style loss than the other two blocks, but produces an overall worse result than the fourth block that manages to maintain better readability.  
\subsection{MSG Net} 
MSG Net was introduced in \cite{zhang2017multi}.We finetuned it to our data that we scraped from the internet: 19 corporate logos (content) and 11 heavy metal logos (style). Style loss hyperparameter was set to 10000, content loss hyperparameter to 1, learning rate to 1.0.\\

\noindent Although MSG Net is more advanced than plain VGG16: it has a fully convolutional architecture, learns weights to evolve an image with the transferred style and has more loss functions, it performs worse than Network of Steel in terms of sparse style transfer, as it does not transfer any font style from heavy metal logos onto the font in the corporate logos at all. MSG-Net manages to evolve some small elements around the glyphs, that are barely noticeable. 
\section{\uppercase{Conclusions}}
\label{sec:conclusion}
\noindent Sparse style transfer requires an approach different to that of other neural style transfer problems, due to a large number of artefacts, merging and distortion of elements of the style and font.\\ 

\noindent In this publication we introduced Network of Steel for sparse style transfer from heavy metal band to corporate logos. We showed that in order to synthesize a readable logo with heavy metal style elements, instead of using layers from all blocks of VGG16, only one or two coarse layers and two or three deep layers are enough. Our future work includes the following challenges:
\begin{enumerate}
    \item Train a separate network for loss coefficients,
    \item Build a large database for training Networks of Steel for different heavy metal styles and corporate logos,
    \item Design accuracy metrics applicable to this problem to enable visual comparison,
    \item In this paper we only used a white background for heavy metal logos, which causes a lot of artefacts. In the future we will use different, more challenging backgrounds, like album covers. 
\end{enumerate}
\noindent We showed that \texttt{conv1${}\_{}$2} is essential to maintaining artefact-free background and layers from the third block in VGG16 learn style faster than deeper layers like \texttt{conv5${}\_{}$3} and \texttt{conv4${}\_{}$3}. Our approach is simple and more robust than \cite{gatys2016image} and \cite{zhang2017multi} for sparse style transfer. The whole deep fourth block (\texttt{conv4${}\_{}$1}, \texttt{conv4${}\_{}$2}, \texttt{conv4${}\_{}$3}) with loss coefficients of $200$ and two coarse layers (\texttt{conv1${}\_{}$1} and \texttt{conv1${}\_{}$2}) with loss coefficients of $2000$ produce the best tradeoff between heavy metal style and the readability of the corporate logo.   

\bibliographystyle{apalike}
{\small
\bibliography{mgn}}

\section*{\uppercase{Appendix}}

\noindent Here we present some of the results that demonstrate the effect of different layer loss coefficients for logo style transfer. In each experiment, all excluded (inactive) layer loss coefficients are set to $0$. All experiments were run for 50000 iterations with the same learning rate and content loss weight of 1, network gradients switched off. Due to the differences in the architecture and training, MSG-Net was evaluated on a number of heavy metal and corporate logos, the best results presented in \ref{tab:msgmodel}.  
\begin{table*}
\caption{Results for VGG16 model, layers \texttt{conv1${}\_{}$2}, \texttt{conv2${}\_{}$2}, \texttt{conv3${}\_{}$3}, \texttt{conv4${}\_{}$3}, \texttt{conv5${}\_{}$3}, as defined in \cite{gatys2016image}}
\label{tab:gatysmodel} 
\centering
\begin{tabular}{|c|c|c|}
  \hline 
  \texttt{c$_{1\_2}$}=\texttt{c$_{2\_2}$}=\texttt{c$_{3\_3}$}=\texttt{c$_{4\_3}$}=\texttt{c$_{5\_3}$}=20&\texttt{c$_{1\_2}$}=\texttt{c$_{2\_2}$}=\texttt{c$_{3\_3}$}=\texttt{c$_{4\_3}$}=\texttt{c$_{5\_3}$}=200&\texttt{c$_{1\_2}$}=\texttt{c$_{2\_2}$}=\texttt{c$_{3\_3}$}=\texttt{c$_{4\_3}$}=\texttt{c$_{5\_3}$}=2000\\
  \hline 
  {\epsfig{file =
  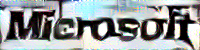, width = 4.0cm}}&{\epsfig{file =
  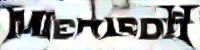, width = 4.0cm}}&{\epsfig{file =
  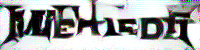, width = 4.0cm}}\\
  \hline
\end{tabular}
\end{table*}
\begin{table*}
\caption{Results for layers \texttt{conv1${}\_{}$2} and \texttt{conv5${}\_{}$3}}
\label{tab:c12c53} 
\centering
\begin{tabular}{|c|c|c|c|}
  \hline
  &\texttt{c$_{5\_3}$}=20&\texttt{c$_{5\_3}$}=200&\texttt{c$_{5\_3}$}=2000\\
  \hline 
  \texttt{c$_{1\_2}$}=0&{\epsfig{file =
  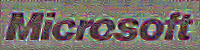, width = 4.0cm}}&{\epsfig{file =
  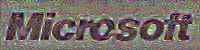, width = 4.0cm}}&{\epsfig{file =
  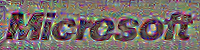, width = 4.0cm}}\\
  \hline
  \texttt{c$_{1\_2}$}=20&{\epsfig{file =
  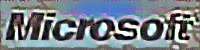, width = 4.0cm}}&{\epsfig{file =
  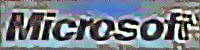, width = 4.0cm}}&{\epsfig{file =
  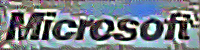, width = 4.0cm}}\\
  \hline
  \texttt{c$_{1\_2}$}=200&{\epsfig{file =
  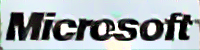, width = 4.0cm}}&{\epsfig{file =
  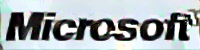, width = 4.0cm}}&{\epsfig{file =
  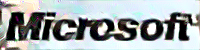, width = 4.0cm}}\\
  \hline
   \texttt{c$_{1\_2}$}=2000&{\epsfig{file =
  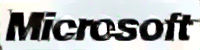, width = 4.0cm}}&{\epsfig{file =
  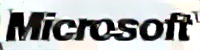, width = 4.0cm}}&{\epsfig{file =
  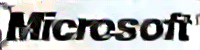, width = 4.0cm}}\\
  \hline
\end{tabular}
\end{table*}

\begin{table*}
\caption{Results for layers \texttt{conv1${}\_{}$2} and \texttt{conv4${}\_{}$3}}
\label{tab:c12c43} 
\centering
\begin{tabular}{|c|c|c|c|}
  \hline
  &\texttt{c$_{4\_3}$}=20&\texttt{c$_{4\_3}$}=200&\texttt{c$_{4\_3}$}=2000\\
  \hline 
  \texttt{c$_{1\_2}$}=0&{\epsfig{file =
  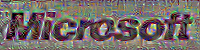, width = 4.0cm}}&{\epsfig{file =
  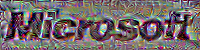, width = 4.0cm}}&{\epsfig{file =
  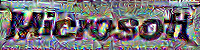, width = 4.0cm}}\\
  \hline
  \texttt{c$_{1\_2}$}=20&{\epsfig{file =
  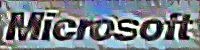, width = 4.0cm}}&{\epsfig{file =
  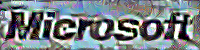, width = 4.0cm}}&{\epsfig{file =
  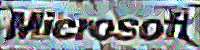, width = 4.0cm}}\\
  \hline
  \texttt{c$_{1\_2}$}=200&{\epsfig{file =
  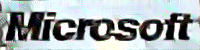, width = 4.0cm}}&{\epsfig{file =
  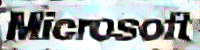, width = 4.0cm}}&{\epsfig{file =
  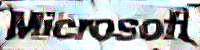, width = 4.0cm}}\\
  \hline
   \texttt{c$_{1\_2}$}=2000&{\epsfig{file =
  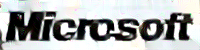, width = 4.0cm}}&{\epsfig{file =
  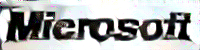, width = 4.0cm}}&{\epsfig{file =
  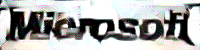, width = 4.0cm}}\\
  \hline
\end{tabular}
\end{table*}

\begin{table*}
\caption{Results for layers \texttt{conv1${}\_{}$2} and \texttt{conv3${}\_{}$3}}
\label{tab:c12c33} 
\centering
\begin{tabular}{|c|c|c|c|}
  \hline
  &\texttt{c$_{3\_3}$}=20&\texttt{c$_{3\_3}$}=200&\texttt{c$_{3\_3}$}=2000\\
  \hline 
  \texttt{c$_{1\_2}$}=0&{\epsfig{file =
  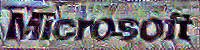, width = 4.0cm}}&{\epsfig{file =
  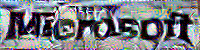, width = 4.0cm}}&{\epsfig{file =
  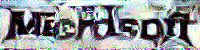, width = 4.0cm}}\\
  \hline
  \texttt{c$_{1\_2}$}=20&{\epsfig{file =
  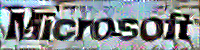, width = 4.0cm}}&{\epsfig{file =
  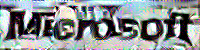, width = 4.0cm}}&{\epsfig{file =
  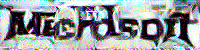, width = 4.0cm}}\\
  \hline
  \texttt{c$_{1\_2}$}=200&{\epsfig{file =
  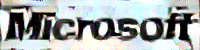, width = 4.0cm}}&{\epsfig{file =
  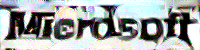, width = 4.0cm}}&{\epsfig{file =
  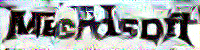, width = 4.0cm}}\\
  \hline
   \texttt{c$_{1\_2}$}=2000&{\epsfig{file =
  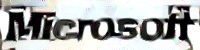, width = 4.0cm}}&{\epsfig{file =
  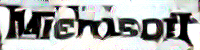, width = 4.0cm}}&{\epsfig{file =
  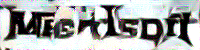, width = 4.0cm}}\\
  \hline
\end{tabular}
\end{table*}

\begin{table*}
\caption{Results for layers \texttt{conv1${}\_{}$2} and \texttt{conv5${}\_{}$2}, \texttt{conv5${}\_{}$3}}
\label{tab:c12c52c53} 
\centering
\begin{tabular}{|c|c|c|c|}
  \hline
  &\texttt{c$_{5\_2}$}=\texttt{c$_{5\_3}$}=20&\texttt{c$_{5\_2}$}=\texttt{c$_{5\_3}$}=200&\texttt{c$_{5\_2}$}=\texttt{c$_{5\_3}$}=2000\\
  \hline 
  \texttt{c$_{1\_2}$}=0&{\epsfig{file =
  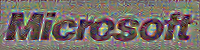, width = 4.0cm}}&{\epsfig{file =
  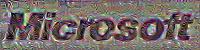, width = 4.0cm}}&{\epsfig{file =
  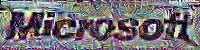, width = 4.0cm}}\\
  \hline
  \texttt{c$_{1\_2}$}=20&{\epsfig{file =
  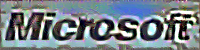, width = 4.0cm}}&{\epsfig{file =
  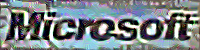, width = 4.0cm}}&{\epsfig{file =
  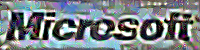, width = 4.0cm}}\\
  \hline
  \texttt{c$_{1\_2}$}=200&{\epsfig{file =
  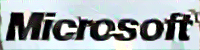, width = 4.0cm}}&{\epsfig{file =
  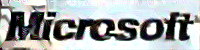, width = 4.0cm}}&{\epsfig{file =
  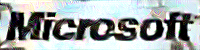, width = 4.0cm}}\\
  \hline
   \texttt{c$_{1\_2}$}=2000&{\epsfig{file =
  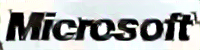, width = 4.0cm}}&{\epsfig{file =
  Images/megadeth_microsoft_49000_steel_net_predicted_l0_200_l44_2000_batch4_l044_1_content_1_contentinitim.png, width = 4.0cm}}&{\epsfig{file =
  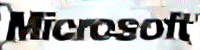, width = 4.0cm}}\\
  \hline
\end{tabular}
\end{table*}
\begin{table*}
\caption{Results for layers \texttt{conv1${}\_{}$2} and \texttt{conv4${}\_{}$2}, \texttt{conv4${}\_{}$3}}
\label{tab:c12c42c43} 
\centering
\begin{tabular}{|c|c|c|c|}
  \hline
  &\texttt{c$_{4\_2}$}=\texttt{c$_{4\_3}$}=20&\texttt{c$_{4\_2}$}=\texttt{c$_{4\_3}$}=200&\texttt{c$_{4\_2}$}=\texttt{c$_{4\_3}$}=2000\\
  \hline 
  \texttt{c$_{1\_2}$}=0&{\epsfig{file =
  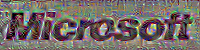, width = 4.0cm}}&{\epsfig{file =
  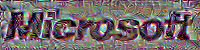, width = 4.0cm}}&{\epsfig{file =
  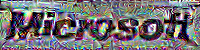, width = 4.0cm}}\\
  \hline
  \texttt{c$_{1\_2}$}=20&{\epsfig{file =
  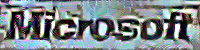, width = 4.0cm}}&{\epsfig{file =
  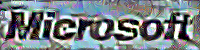, width = 4.0cm}}&{\epsfig{file =
  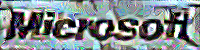, width = 4.0cm}}\\
  \hline
  \texttt{c$_{1\_2}$}=200&{\epsfig{file =
  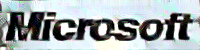, width = 4.0cm}}&{\epsfig{file =
  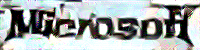, width = 4.0cm}}&{\epsfig{file =
  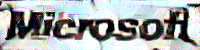, width = 4.0cm}}\\
  \hline
   \texttt{c$_{1\_2}$}=2000&{\epsfig{file =
  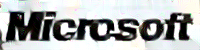, width = 4.0cm}}&{\epsfig{file =
  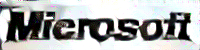, width = 4.0cm}}&{\epsfig{file =
  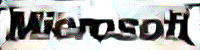, width = 4.0cm}}\\
  \hline
\end{tabular}
\end{table*}
\begin{table*}
\caption{Results for layers \texttt{conv1${}\_{}$2} and \texttt{conv3${}\_{}$2}, \texttt{conv3${}\_{}$3}}
\label{tab:c12c32c33} 
\centering
\begin{tabular}{|c|c|c|c|}
  \hline
  &\texttt{c$_{3\_2}$}=\texttt{c$_{3\_3}$}=20&\texttt{c$_{3\_2}$}=\texttt{c$_{3\_3}$}=200&\texttt{c$_{3\_2}$}=\texttt{c$_{3\_3}$}=2000\\
  \hline 
  \texttt{c$_{1\_2}$}=0&{\epsfig{file =
  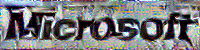, width = 4.0cm}}&{\epsfig{file =
  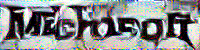, width = 4.0cm}}&{\epsfig{file =
  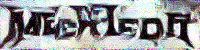, width = 4.0cm}}\\
  \hline
  \texttt{c$_{1\_2}$}=20&{\epsfig{file =
  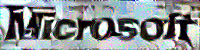, width = 4.0cm}}&{\epsfig{file =
  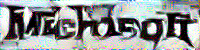, width = 4.0cm}}&{\epsfig{file =
  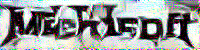, width = 4.0cm}}\\
  \hline
  \texttt{c$_{1\_2}$}=200&{\epsfig{file =
  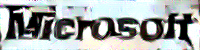, width = 4.0cm}}&{\epsfig{file =
  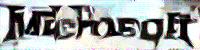, width = 4.0cm}}&{\epsfig{file =
  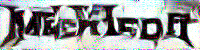, width = 4.0cm}}\\
  \hline
   \texttt{c$_{1\_2}$}=2000&{\epsfig{file =
  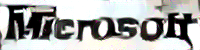, width = 4.0cm}}&{\epsfig{file =
  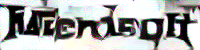, width = 4.0cm}}&{\epsfig{file =
  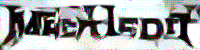, width = 4.0cm}}\\
  \hline
\end{tabular}
\end{table*}
\begin{table*}
\caption{Results for the first and the fifth block: \texttt{conv1${}\_{}$1} \texttt{conv1${}\_{}$2}, and \texttt{conv5${}\_{}$1}, \texttt{conv5${}\_{}$2}, \texttt{conv5${}\_{}$3}}
\label{tab:c11c12c51c52c53} 
\centering
\begin{tabular}{|c|c|c|c|}
  \hline
  &\texttt{c$_{5\_1}$}=\texttt{c$_{5\_2}$}=\texttt{c$_{5\_3}$}=20&\texttt{c$_{5\_1}$}=\texttt{c$_{5\_2}$}=\texttt{c$_{5\_3}$}=200&\texttt{c$_{5\_1}$}=\texttt{c$_{5\_2}$}=\texttt{c$_{5\_3}$}=2000\\
  \hline 
  \texttt{c$_{1\_1}$=c$_{1\_2}$}=0&{\epsfig{file =
  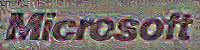, width = 4.0cm}}&{\epsfig{file =
  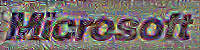, width = 4.0cm}}&{\epsfig{file =
  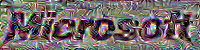, width = 4.0cm}}\\
  \hline
  \texttt{c$_{1\_1}$=c$_{1\_2}$}=20&{\epsfig{file =
  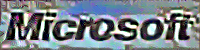, width = 4.0cm}}&{\epsfig{file =
  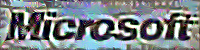, width = 4.0cm}}&{\epsfig{file =
  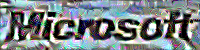, width = 4.0cm}}\\
  \hline
  \texttt{c$_{1\_1}$=c$_{1\_2}$}=200&{\epsfig{file =
  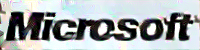, width = 4.0cm}}&{\epsfig{file =
  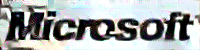, width = 4.0cm}}&{\epsfig{file =
  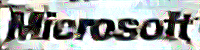, width = 4.0cm}}\\
  \hline
   \texttt{c$_{1\_1}$=c$_{1\_2}$}=2000&{\epsfig{file =
  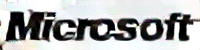, width = 4.0cm}}&{\epsfig{file =
  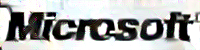, width = 4.0cm}}&{\epsfig{file =
  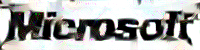, width = 4.0cm}}\\
  \hline
\end{tabular}
\end{table*}
\begin{table*}
\caption{Results for the first and the fourth block: \texttt{conv1${}\_{}$1} \texttt{conv1${}\_{}$2}, and \texttt{conv4${}\_{}$1}, \texttt{conv4${}\_{}$2}, \texttt{conv4${}\_{}$3}}
\label{tab:c11c12c41c42c43} 
\centering
\begin{tabular}{|c|c|c|c|}
  \hline
  &\texttt{c$_{4\_1}$}=\texttt{c$_{4\_2}$}=\texttt{c$_{4\_3}$}=20&\texttt{c$_{4\_1}$}=\texttt{c$_{4\_2}$}=\texttt{c$_{4\_3}$}=200&\texttt{c$_{4\_1}$}=\texttt{c$_{4\_2}$}=\texttt{c$_{4\_3}$}=2000\\
  \hline 
  \texttt{c$_{1\_1}$=c$_{1\_2}$}=0&{\epsfig{file =
  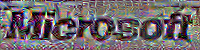, width = 4.0cm}}&{\epsfig{file =
  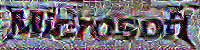, width = 4.0cm}}&{\epsfig{file =
  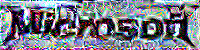, width = 4.0cm}}\\
  \hline
  \texttt{c$_{1\_1}$=c$_{1\_2}$}=20&{\epsfig{file =
  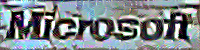, width = 4.0cm}}&{\epsfig{file =
  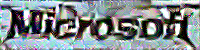, width = 4.0cm}}&{\epsfig{file =
  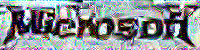, width = 4.0cm}}\\
  \hline
  \texttt{c$_{1\_1}$=c$_{1\_2}$}=200&{\epsfig{file =
  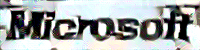, width = 4.0cm}}&{\epsfig{file =
  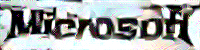, width = 4.0cm}}&{\epsfig{file =  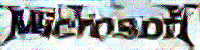, width = 4.0cm}}\\
  \hline
   \texttt{c$_{1\_1}$=c$_{1\_2}$}=2000&{\epsfig{file =
  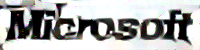, width = 4.0cm}}&{\epsfig{file =
  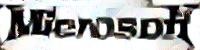, width = 4.0cm}}&{\epsfig{file =
  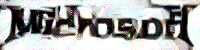, width = 4.0cm}}\\
  \hline
\end{tabular}
\end{table*}
\begin{table*}
\caption{Results for the first and the third block: \texttt{conv1${}\_{}$1} \texttt{conv1${}\_{}$2}, and \texttt{conv3${}\_{}$1}, \texttt{conv3${}\_{}$2}, \texttt{conv3${}\_{}$3}}
\label{tab:c11c12c31c31c33} 
\centering
\begin{tabular}{|c|c|c|c|}
  \hline
  &\texttt{c$_{3\_1}$}=\texttt{c$_{3\_2}$}=\texttt{c$_{3\_3}$}=20&\texttt{c$_{3\_1}$}=\texttt{c$_{3\_2}$}=\texttt{c$_{3\_3}$}=200&\texttt{c$_{3\_1}$}=\texttt{c$_{3\_2}$}=\texttt{c$_{3\_3}$}=2000\\
  \hline 
  \texttt{c$_{1\_1}$=c$_{1\_2}$}=0&{\epsfig{file =
  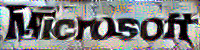, width = 4.0cm}}&{\epsfig{file =
  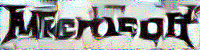, width = 4.0cm}}&{\epsfig{file =
  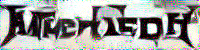, width = 4.0cm}}\\
  \hline
  \texttt{c$_{1\_1}$=c$_{1\_2}$}=20&{\epsfig{file =
  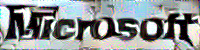, width = 4.0cm}}&{\epsfig{file =
  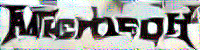, width = 4.0cm}}&{\epsfig{file =
  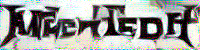, width = 4.0cm}}\\
  \hline
  \texttt{c$_{1\_1}$=c$_{1\_2}$}=200&{\epsfig{file =
  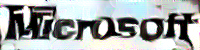, width = 4.0cm}}&{\epsfig{file =
  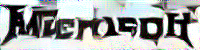, width = 4.0cm}}&{\epsfig{file =
  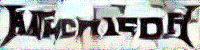, width = 4.0cm}}\\
  \hline
   \texttt{c$_{1\_1}$=c$_{1\_2}$}=2000&{\epsfig{file =
  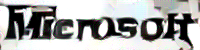, width = 4.0cm}}&{\epsfig{file =
  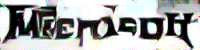, width = 4.0cm}}&{\epsfig{file =
  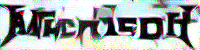, width = 4.0cm}}\\
  \hline
\end{tabular}
\end{table*}

\vspace{-6cm}
\begin{table*}
\caption{Results for MSG-Net model}
\label{tab:msgmodel} 
\centering
\begin{tabular}{|c|c|c|c|}
   \hline
   &{\epsfig{file =
  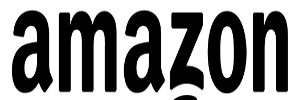, width = 4.0cm, height = 1.5cm}}&{\epsfig{file =
  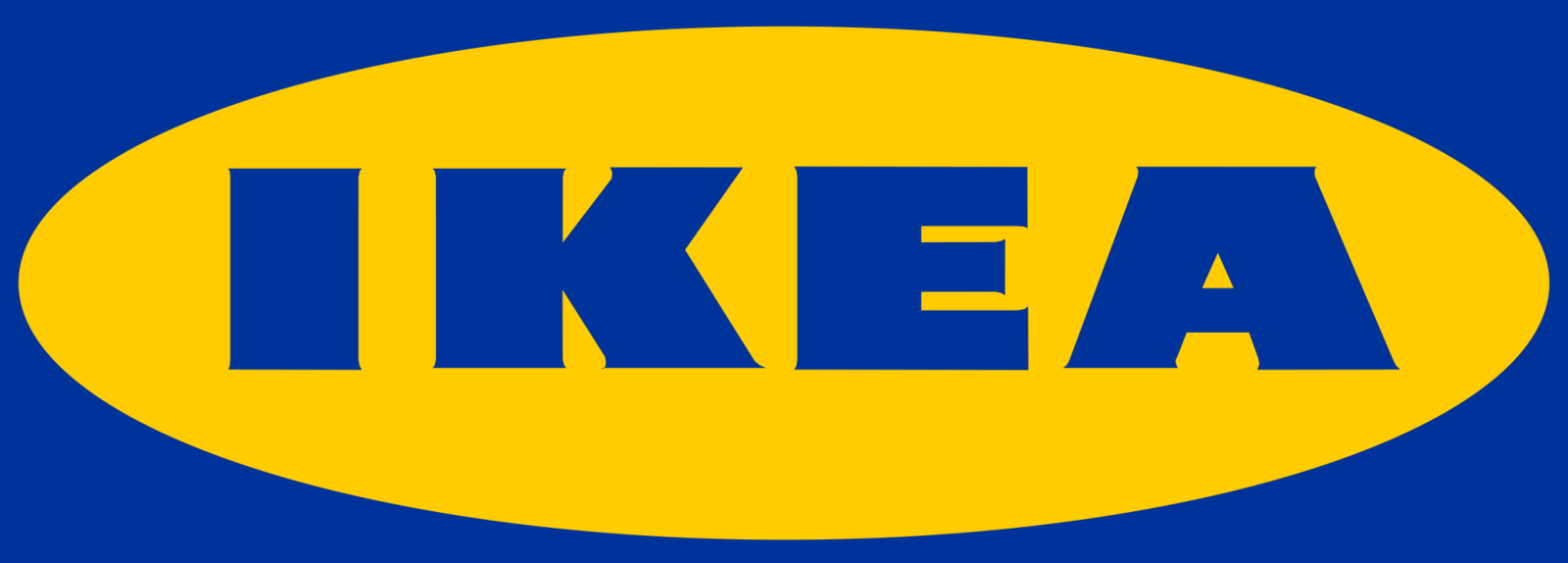, width = 4.0cm, height = 1.5cm}}&{\epsfig{file =
  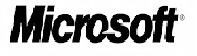, width = 4.0cm, height = 1.5cm}}\\    
  \hline 
  {\epsfig{file =
  Images/megadeth_800_200.jpg, width = 4.0cm, height = 1.5cm}}&{\epsfig{file =
  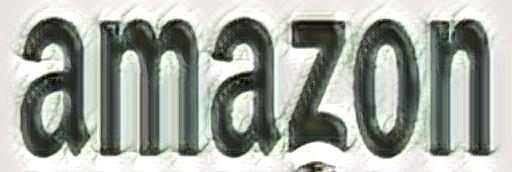, width = 4.0cm, height = 1.5cm}}&{\epsfig{file =
  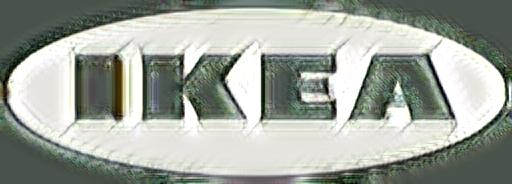, width = 4.0cm, height = 1.5cm}}&{\epsfig{file =
  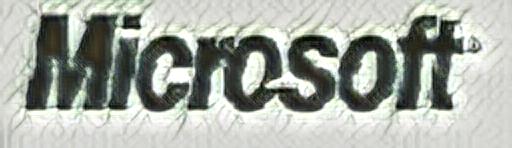, width = 4.0cm, height = 1.5cm}}\\
  \hline 
  {\epsfig{file =
  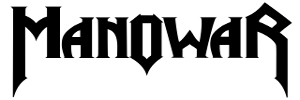, width = 4.0cm}}&{\epsfig{file =
  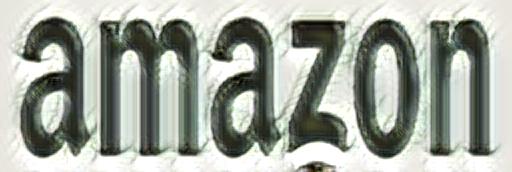, width = 4.0cm, height = 1.5cm}}&{\epsfig{file =
  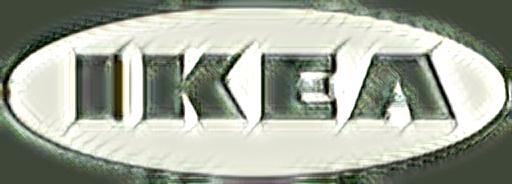, width = 4.0cm, height = 1.5cm}}&{\epsfig{file =
  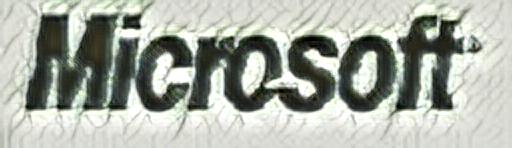, width = 4.0cm, height = 1.5cm}}\\
  \hline
  {\epsfig{file =
  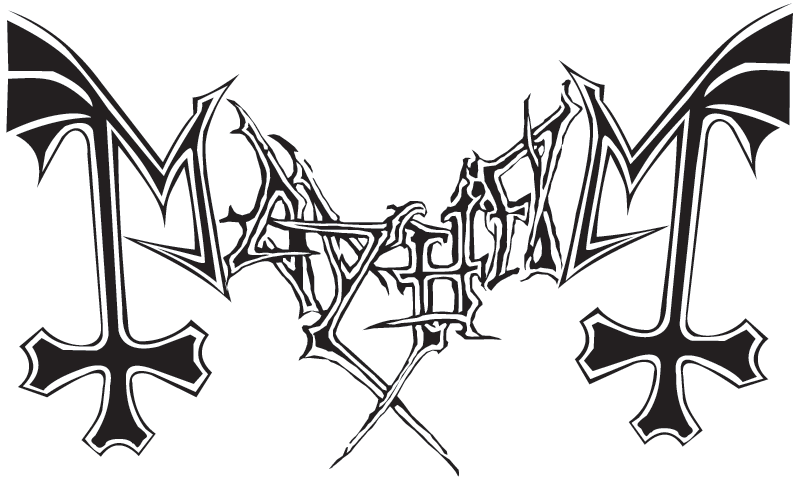, width = 4.0cm, height = 2.0cm}}&{\epsfig{file =
  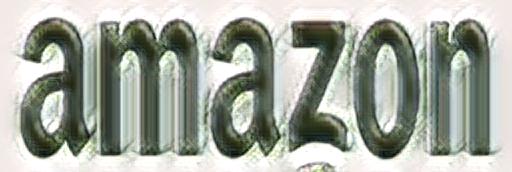, width = 4.0cm, height = 1.5cm}}&{\epsfig{file =
  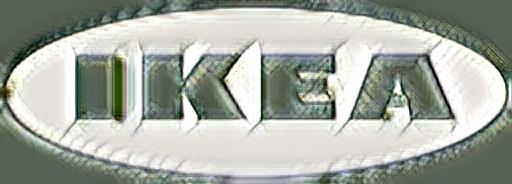, width = 4.0cm, height = 1.5cm}}&{\epsfig{file =
  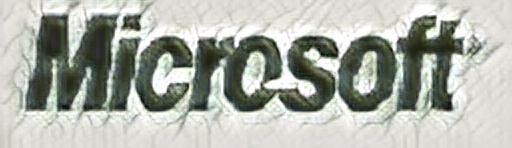, width = 4.0cm, height = 1.5cm}}\\
  \hline
  {\epsfig{file =
  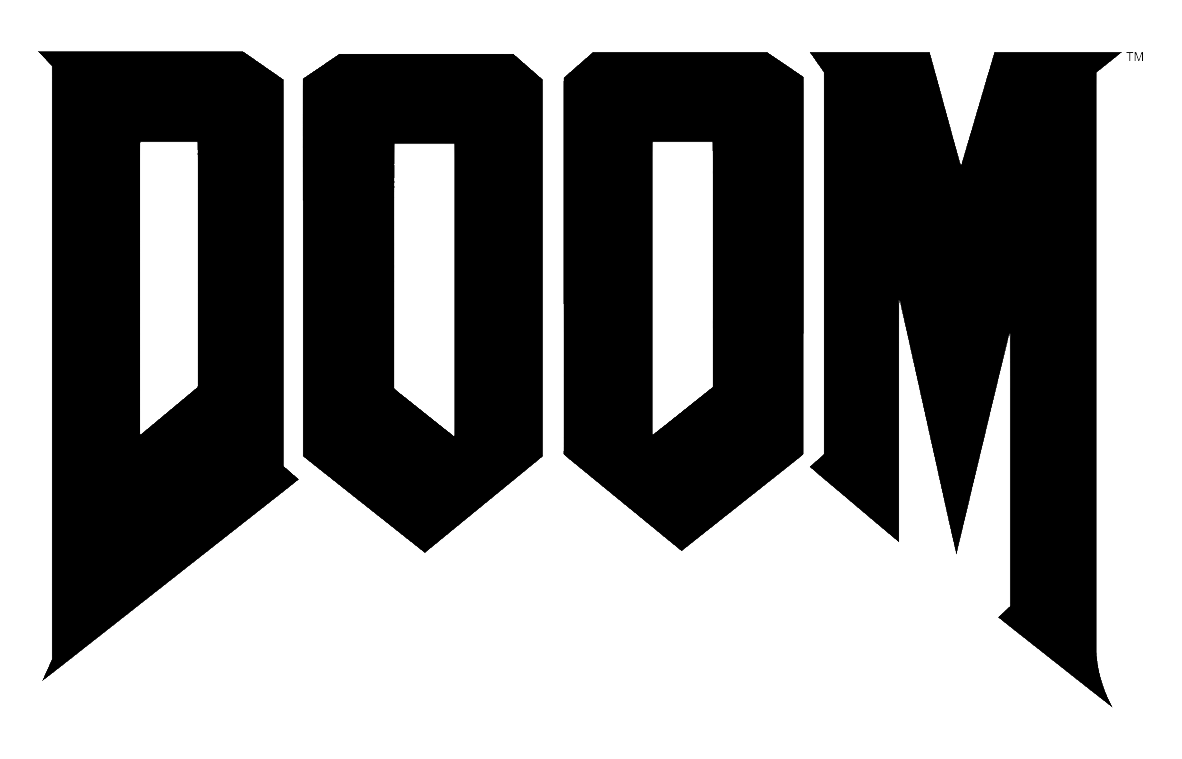, width = 4.0cm,  height = 2.0cm}}&{\epsfig{file =
  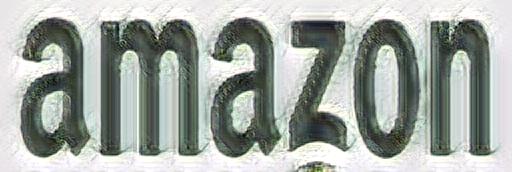, width = 4.0cm, height = 1.5cm}}&{\epsfig{file =
  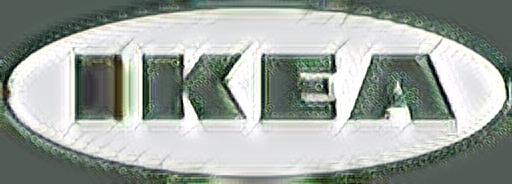, width = 4.0cm, height = 1.5cm}}&{\epsfig{file =
  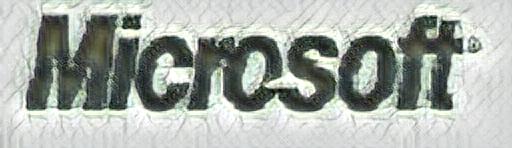, width = 4.0cm, height = 1.5cm}}\\
  \hline
  {\epsfig{file =
  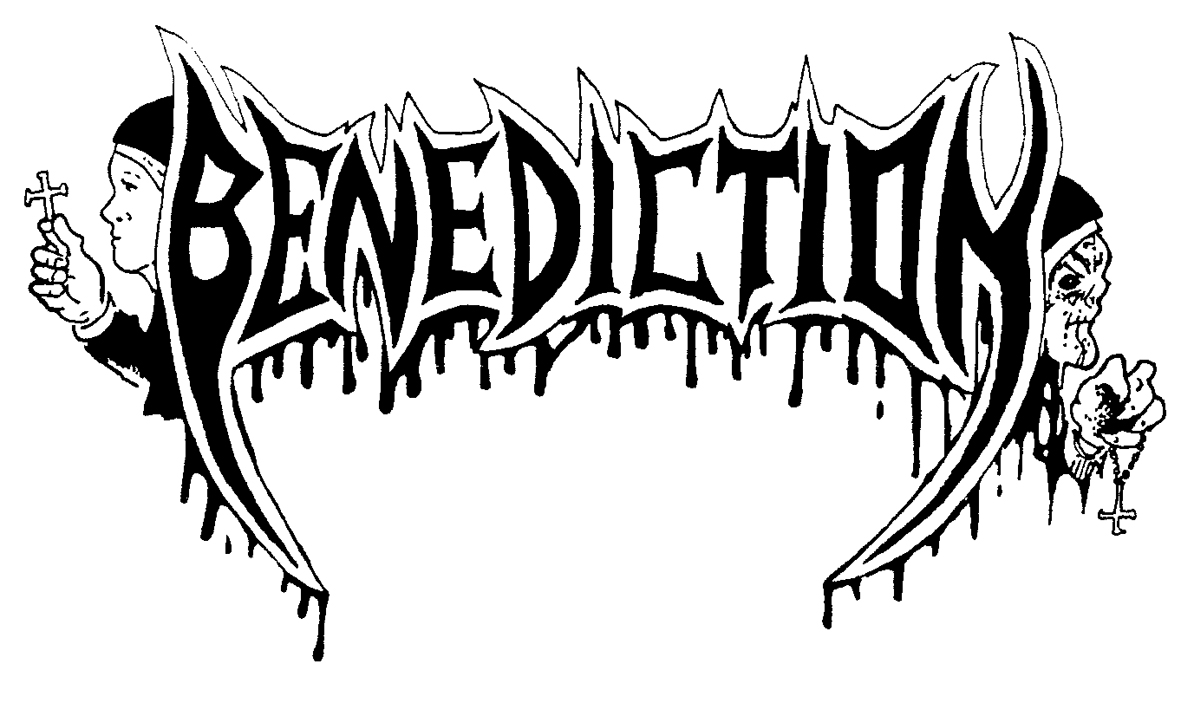, width = 4.0cm, height = 2.0cm}}&{\epsfig{file =
  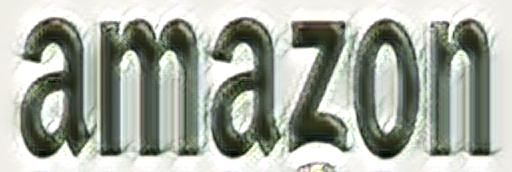, width = 4.0cm, height = 1.5cm}}&{\epsfig{file =
  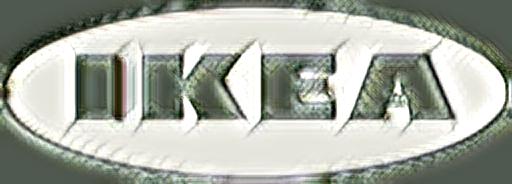, width = 4.0cm, height = 1.5cm}}&{\epsfig{file =
  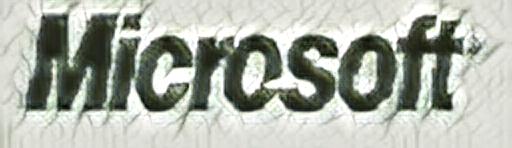, width = 4.0cm, height = 1.5cm}}\\
  \hline
\end{tabular}
\end{table*}
\vspace{-6cm}
\end{document}